\documentclass[conference]{IEEEtran}
\usepackage{cite}
\usepackage{amsmath,amssymb,amsfonts}
\usepackage{algorithmic}
\usepackage{graphicx}
\usepackage{textcomp}
\usepackage{xcolor}
\def\BibTeX{{\rm B\kern-.05em{\sc i\kern-.025em b}\kern-.08em
    T\kern-.1667em\lower.7ex\hbox{E}\kern-.125emX}}
\usepackage{graphicx}

\usepackage{pgfplots}
\usepackage{pgfplotstable}
\usepackage{booktabs, colortbl, array}
\usepackage{enumitem}
\usepackage{rotating}
\usepackage{balance}
\usepackage{bbm}
\usepackage{amssymb}
\usepackage{amsmath}
\usepackage{amsthm}

\usepackage{url}

\begin{document}
\title{Algoritmos de minería de datos en la industria sanitaria}
%
%
\author{\IEEEauthorblockN{Marta Li Wang}
\IEEEauthorblockA{\textit{Wufong University, China}
}}
\maketitle              
\begin{abstract}
En este estudio se examinan varios enfoques de la minería de datos (DM) en la industria de la salud procedentes de muchos grupos de investigación de todo el mundo. La atención se centra en los modernos procesadores multinúcleo incorporados a los ordenadores básicos actuales, que suelen encontrarse en los institutos universitarios tanto en forma de pequeños servidores como de estaciones de trabajo. Por tanto, no son deliberadamente ordenadores de alto rendimiento. Los procesadores multinúcleo modernos constan de varios (de 2 a más de 100) núcleos informáticos, que funcionan de forma independiente entre sí según el principio de ``instrucciones múltiples, datos múltiples'' (MIMD). Tienen una memoria principal común (memoria compartida). Cada uno de estos núcleos informáticos dispone de varias (2-16) unidades aritméticas-lógicas, que pueden realizar simultáneamente la misma operación aritmética sobre varios datos de forma vectorial (single instruction multiple data, SIMD). Los algoritmos de DM deben utilizar ambos tipos de paralelismo (SIMD y MIMD), siendo el acceso a la memoria principal (componente centralizado) la principal barrera para aumentar la eficiencia. Esto es importante para la DM en aplicaciones de la industria de la salud como ECG, EEG, CT, SPECT, fMRI, DTI, ultrasonido, microscopía, dermascopía, etc.

\textit{Abstract (ingles)---}In this paper, we review data mining approaches for health applications. Our focus is on hardware-centric approaches. Modern computers consist of multiple processors, each equipped with multiple cores, each with a set of arithmetic/logical units. Thus, a modern computer may be composed of several thousand units capable of doing arithmetic operations like addition and multiplication. Graphic processors, in addition may offer some thousand such units. In both cases, single instruction multiple data and multiple instruction multiple data parallelism must be exploited. We review the principles of algorithms which exploit this parallelism and focus also on the memory issues when multiple processing units access main memory through caches. This is important for many applications of health, such as ECG, EEG, CT, SPECT, fMRI, DTI, ultrasound, microscopy, dermascopy, etc.

\end{abstract}

\section{Introducción}
Las aplicaciones de la industria de la salud, como ECG, EEG, CT, SPECT, fMRI, DTI, ultrasonido, microscopía, dermascopía, etc., plantean elevados requisitos en cuanto al rendimiento en tiempo de ejecución de la minería de datos (DM).
El hardware informático moderno permite el desarrollo de aplicaciones de alto rendimiento para el análisis de datos en muchos niveles diferentes. La atención se centra en los modernos procesadores multinúcleo incorporados a los ordenadores básicos actuales, que suelen encontrarse en los institutos universitarios tanto en forma de pequeños servidores como de estaciones de trabajo. Por tanto, no son deliberadamente ordenadores de alto rendimiento. Los procesadores multinúcleo modernos constan de varios (de 2 a más de 100) núcleos informáticos, que funcionan de forma independiente entre sí según el principio de ``instrucciones múltiples, datos múltiples'' (MIMD). Tienen una memoria principal común (memoria compartida). Cada uno de estos núcleos informáticos dispone de varias (2-16) unidades aritméticas-lógicas, que pueden realizar simultáneamente la misma operación aritmética sobre varios datos de forma vectorial (single instruction multiple data, SIMD). Los algoritmos de DM deben utilizar ambos tipos de paralelismo (SIMD y MIMD), siendo el acceso a la memoria principal (componente centralizado) la principal barrera para aumentar la eficiencia. En principio, las cachés rápidas permiten resolver este problema. Sin embargo, los algoritmos de alto rendimiento deben diseñarse de forma que estas cachés puedan funcionar de forma eficaz (con conciencia de caché o sin ella).

Para las tareas estándar, como el álgebra lineal, la optimización o el entrenamiento de redes neuronales, ya se utilizan ampliamente algoritmos de alto rendimiento con paralelismo SIMD y MIMD (en parte con un uso eficiente de la caché). Los métodos más especializados, como el análisis de clústeres, requieren un nuevo desarrollo específico de los algoritmos, como se ha publicado en numerosos artículos sobre los métodos de clúster k-Means, EM y DBSCAN y otros algoritmos para la GPU y las arquitecturas multinúcleo. Se trata sobre todo del desarrollo de principios algorítmicos generales que pueden aplicarse a muchos problemas, por ejemplo, propuestas de bucles oblicuos a la caché. La idea de los bucles olvidados por la caché es reemplazar los bucles anidados (que desempeñan un papel central en muchos algoritmos de aplicación) de manera que (1) puedan distribuirse a las unidades paralelas de forma que se preserve la localidad y (2) los propios accesos se diseñen de forma que se consiga un alto grado de localidad y, por tanto, un uso eficiente de la caché. Esto se consigue utilizando una variante especial de la curva de Hilbert (una curva que llena el espacio llamada FUR-Hilbert). En comparación con la versión básica de los algoritmos (sin paralelismo SIMD y MIMD), a menudo es posible tener factores de aceleración de dos o más órdenes de magnitud (es decir, más de 100).

El objetivo general de esta área de investigación es seguir desarrollando estas técnicas de paralelización para algoritmos de análisis de datos, así como su uso ejemplar en las aplicaciones punteras de la sanidad y las ciencias de la vida.

Un objetivo interesante sería seguir desarrollando el paradigma algorítmico básico de definición de algoritmos de alto rendimiento para el análisis de datos utilizando el concepto de bucles oblicuos de caché con los siguientes objetivos
\begin{itemize}
\item El rendimiento se incrementará aún más en comparación con el enfoque anterior, por ejemplo, mediante el equilibrio de carga dinámico.
\item Se mejorará la usabilidad para varios algoritmos de aplicación nuevos, por ejemplo, mediante el desarrollo de algoritmos especializados para detectar dependencias de datos y desarrollar criterios de convergencia para dichos algoritmos
\item El concepto de bucles sin caché se desarrollará para otras arquitecturas, por ejemplo, GPU o sistemas distribuidos (por ejemplo, entornos de red y nube).
\end{itemize}

\section{Los bucles sin caché con equilibrio de carga dinámico}
Un enfoque de paralelización basado en SIMD y MIMD para diseñar pares de bucles de manera que soporten de forma flexible cachés de cualquier tamaño divide la carga de trabajo total de un algoritmo mediante las llamadas curvas de llenado de espacio en paquetes que se asignan a los núcleos individuales y hacen un uso eficiente de la memoria caché en ellos. Aunque la carga de trabajo puede distribuirse inicialmente de forma óptima entre los núcleos del ordenador de esta manera tan sencilla, todavía no es posible reequilibrar dinámicamente la carga si, por ejemplo, los núcleos están sometidos a una carga básica desigual debido a influencias externas. En este paquete de trabajo, los procedimientos para la redistribución dinámica de la carga deben diseñarse de forma que no entren en colisión con el objetivo general de un uso eficiente de la caché.

\section{Modelación de estructuras de bucle}
El enfoque actual se limita a atravesar dos bucles anidados con límites de bucle fijos de forma que se utilicen cachés de cualquier tamaño y también se permita el paralelismo MIMD. Para ello se ha utilizado una variante de la curva de Hilbert recientemente desarrollada. En este paquete de trabajo, este concepto se ampliará a cualquier número de bucles y se podrán modelar las dependencias de los bucles entre sí. Esto debería permitir procesar eficazmente, por ejemplo, matrices triangulares o de banda, así como tensores de orden superior. Para el procesamiento de tres o más bucles anidados, ya existen básicamente las extensiones matemáticas de la curva de Hilbert y otras curvas de llenado de espacio en espacios de mayor dimensión. Sin embargo, estos métodos todavía tienen que ser ampliados y adaptados de tal manera que el cálculo se pueda realizar de forma muy eficiente (en tiempo constante por cada paso de bucle), lo que se consiguió en el caso bidimensional con la ayuda de diferentes conceptos. La extensión de estos conceptos para espacios de dimensionalidad arbitraria nos parece una tarea de investigación exigente pero también factible. Para la modelización de las dependencias es posible definir estructuras de datos adecuadas que se orienten y apoyen el carácter jerárquico-recursivo de las curvas de llenado del espacio.

\section{Modelación de los requisitos de monotonía}
Este enfoque, basado en una variante especial de la curva de Hilbert, ofrece la ventaja de que en la mayoría de las aplicaciones el efecto sobre la ubicación de la caché es más fuerte en comparación con las curvas de llenado de espacio comparables. Sin embargo, algunos algoritmos de aplicación requieren una secuencia de pases de bucle que cumpla ciertas propiedades de monotonía, es decir, que ciertos índices de bucle se procesen antes que otros. La idea es combinar las curvas de llenado de espacio existentes, como la de Hilbert o la de orden Z, de manera que incluso los requisitos de monotonía parcial puedan representarse en dimensiones individuales. En este paquete de trabajo, se ampliará el concepto de bucles con memoria caché para especificar las propiedades de monotonicidad y aplicar automáticamente las curvas de llenado de espacio adecuadas.

\section{GPU y Sistemas Distribuidos}
El problema de que los algoritmos trabajen distribuidos sobre los datos de forma que cada unidad de procesamiento alcance la mayor localidad de acceso posible no sólo se plantea en los sistemas de CPU multinúcleo, sino también en otras arquitecturas distribuidas y paralelas. El escenario es muy similar para las unidades de procesamiento gráfico (GPU), pero el problema básico es similar para sistemas distribuidos sin memoria física compartida (grids, cloud computing). En este paquete de trabajo se investigará la extensión a dichas arquitecturas. Aunque el problema básico de la preservación de la localidad sigue siendo el mismo, en escenarios de hardware alternativos las consideraciones sobre el equilibrio entre los diferentes factores de coste de los algoritmos, es decir, especialmente el equilibrio entre los costes de transmisión de los datos a través de la red o de las conexiones de bus internas y el tiempo de computación (que también incluye la gestión de las curvas de llenado de espacio y otras técnicas para la preservación de la localidad) pueden variar en detalle.

\section{Derivación de los criterios de aplicabilidad}
No todos los algoritmos de análisis de datos son igualmente aptos para su paralelización mediante bucles con memoria caché, aunque sigan el patrón básico de bucles anidados. Por lo tanto, en este paquete de trabajo se desarrolla un conjunto de criterios para encontrar dependencias de datos y requisitos de monotonía en los algoritmos de aplicación de forma manual o (semi)automática. Además, se van a desarrollar criterios para determinar cuándo los algoritmos no pueden transformarse de forma equivalente de manera demostrable, pero después de una transformación, resultan algoritmos que también son convergentes y posiblemente alcanzan un óptimo local que se desvía del algoritmo original. Desde el punto de vista científico, este último es el mayor reto, porque ya se han propuesto técnicas automáticas para el reconocimiento de las dependencias de datos, que se utilizan con mucho éxito y de forma generalizada en otras subáreas de la informática (por ejemplo, la construcción de compiladores). En cambio, las técnicas de transformación en algoritmos meramente equivalentes a los resultados son mucho menos conocidas.

\section{Trabajo relacionado}

Hay varios enfoques para \textbf{diferentes tipos de datos}. Los datos pueden ser de cualquier tipo, siempre que exista una función de distancia. Los datos de texto de longitud fija suelen utilizar la distancia de Hamming \cite{DBLP:journals/tjs/HoOK18} y la similitud entre textos de longitud variable suele medirse mediante la distancia de edición \cite{DBLP:journals/pvldb/XiaoWL08}. Una medida común para los datos de conjuntos es la distancia Jaccard \cite{DBLP:conf/sigmod/DengT018,DBLP:journals/tods/XiaoWLYW11}, mientras que la similitud de los documentos se procesa con medidas de similitud tipo coseno \cite{DBLP:series/synthesis/2013Augsten,DBLP:conf/icde/ShangLLF17}.

\textbf{Las técnicas de búsqueda del vecino más cercano} también pueden aplicarse al problema de la unión de similitudes, pero sin garantías de integridad y exactitud del resultado. Puede haber falsos positivos, así como falsos negativos. Recientemente se ha utilizado una aproximación \cite{DBLP:journals/tkde/YuNLWY17} al Locality Sensitive Hashing (LSH) en una muestra de puntos representativa, para reducir el número de operaciones de búsqueda. El LSH es de interés en el trabajo teórico fundacional, donde se propuso un enfoque de LSH recursivo y sin caché \cite{DBLP:journals/algorithmica/PaghPSS17}. El tema de las soluciones aproximadas para la unión de similitudes es también un campo emergente en el aprendizaje profundo \cite{DBLP:journals/corr/abs-1803-04765}. Hay enfoques aproximados que se dirigen a casos de baja dimensión (uniones espaciales en 2--3 dimensiones \cite{DBLP:conf/icde/BryanEF08}) o a casos de mayor dimensión (10-20) \cite{DBLP:conf/focs/AndoniI06}. Los casos de muy alta dimensión, con dimensiones de $128$ y superiores, han sido objeto de técnicas de aproximación simbólica (SAX) \cite{DBLP:journals/concurrency/MaJZ17}) para generar candidatos aproximados. Las técnicas SAX se basan en varios parámetros indirectos como el tamaño del PAA o el tamaño del alfabeto iSAX.

Existen técnicas de indexación preconstruidas, que se basan en \textbf{curvas de llenado de espacio} y se aplican al problema de unión por similitud. En concreto, donde los datos se ordenan de forma eficiente con respecto a una o más curvas de orden Z \cite{DBLP:conf/kdd/DittrichS01, DBLP:journals/tkde/KoudasS00, DBLP:conf/icde/LiebermanSS08} para comprobar la intersección de los hipercubos en las estructuras de datos. Otros proponen curvas de llenado de espacio, para reducir el coste de almacenamiento del índice \cite{DBLP:journals/tkde/ChenGLJC17}.
GESS \cite{DBLP:conf/kdd/DittrichS01} y
LESS \cite{DBLP:conf/icde/LiebermanSS08} se dirigen a las GPU y no a los entornos multinúcleo. ZC y MSJ \cite{DBLP:journals/tkde/KoudasS00} así como el índice SPB-tree \cite{DBLP:journals/tkde/ChenGLJC17}, aunque son sencillos, requieren transformaciones espaciales y preprocesamiento, lo que dificulta su paralelización.

\textbf{familiaEGO} de algoritmos $\epsilon$-join. El algoritmo EGO-join es el primer algoritmo de esta familia introducido por B\"{o}hm et al. en \cite{epsilongridorder}. El Epsilon Grid Order (EGO) fue introducido como un orden estricto (es decir, un orden que es irreflexivo, asimétrico y transitivo). Se demostró que todos los socios de unión de algún punto $\mathbf x$ se encuentran dentro de un intervalo $\epsilon$, del Orden de Rejilla Epsilon. Los algoritmos de la familia EGO explotan este conocimiento para la operación de unión. El EGO-join ha sido reimplementado como una variante recursiva con heurística adicional, para decidir rápidamente si dos secuencias son no unibles \cite{DBLP:conf/dasfaa/KalashnikovP03}. Otras mejoras propusieron dos nuevos miembros de esta familia, el algoritmo EGO$^{*}$ \cite{DBLP:journals/is/KalashnikovP07} y su versión extendida llamada Super-EGO \cite{DBLP:journals/vldb/Kalashnikov13}, que se dirige a entornos multinúcleo utilizando un modelo de programación multiproceso/multihilo. Super-EGO propone una reordenación dimensional \cite{DBLP:journals/vldb/Kalashnikov13}. En los experimentos, Super-EGO encuentra algunas dificultades con los datos distribuidos uniformemente, especialmente cuando el número de objetos de datos supera los millones de puntos o la dimensionalidad es superior a $32$.

Si la unión por similitud se ejecuta varias veces en las mismas instancias de los datos, se podría considerar \textbf{enfoques basados en índices} \cite{DBLP:conf/icde/BohmK01, DBLP:journals/jda/ParedesR09, DBLP:journals/tkde/ChenGLJC17}, como R-tree \cite{DBLP:conf/sigmod/BrinkhoffKS93} o \textit{M}-tree \cite{DBLP:conf/vldb/CiacciaPZ97}. Los enfoques basados en índices tienen el potencial de reducir el tiempo de ejecución, ya que el índice almacena información precomputada que reduce significativamente el tiempo de ejecución de la consulta. Este paso precomputacional podría ser costoso, especialmente en el caso de la Lista de Clusters Gemelos (LTC)
\cite{DBLP:journals/jda/ParedesR09}, donde el algoritmo necesita construir índices conjuntos o combinados para cada par de puntos del conjunto de datos. El D-Index \cite{DBLP:journals/mta/DohnalGSZ03} y sus extensiones (es decir, el eD-Index \cite{DBLP:conf/dexa/DohnalGZ03} o el índice i-Sim \cite{DBLP:conf/sisap/PearsonS14}) construyen una estructura jerárquica de niveles de índice, donde cada nivel se organiza en cubos separables y un conjunto de exclusión. El inconveniente más importante de D-Index, eD-Index e i-Sim es que pueden requerir la reconstrucción de la estructura del índice para diferentes $\epsilon$.
Sin embargo, los enfoques de indexación espacial no funcionan bien en espacios de alta dimensión, debido a la "maldición de la dimensionalidad".

\textbf{La partición de datos en múltiples máquinas} no es el objetivo principal de este trabajo, en el que asumimos que los datos caben en la memoria principal. El caso de los algoritmos de unión relacional se ha estudiado ampliamente en el pasado \cite{DBLP:conf/sigmod/SchneiderD89, DBLP:conf/kdd/WangMP13, DBLP:journals/pvldb/FierABLF18}.
La unión por similitud se ha aplicado con éxito en el entorno distribuido con diferentes variantes de MapReduce \cite{DBLP:conf/waim/LiWU16, DBLP:conf/sigmod/McCauley018, DBLP:journals/pvldb/FierABLF18}. Otra versión distribuida se propone en \cite{DBLP:conf/sigmod/ZhaoRDW16}. Allí se utiliza una solución multinodo con balanceo de carga, que no requiere re-partición en los datos de entrada. Esta variante se centra en la minimización de la transferencia de datos, la congestión de la red y el equilibrio de carga entre múltiples nodos.

La unión por similitud ya ha sido implementada para \textbf{Graphics Processing Units (GPUs)}.
En \cite{DBLP:conf/btw/BohmNPZ09} los autores utilizan una estructura de directorios para generar puntos candidatos. En conjuntos de datos con 8 millones de puntos, el algoritmo propuesto para la GPU es más rápido que su variante para la CPU, cuando el
 $\epsilon$-región tiene al menos 1 o 2 vecinos medios.
LSS \cite{DBLP:conf/icde/LiebermanSS08} es otra variante de similaridad para la GPU, que es adecuada para datos de alta dimensión. Lamentablemente, tanto \cite{DBLP:conf/icde/LiebermanSS08} como \cite{DBLP:conf/btw/BohmNPZ09} están dirigidos a las GPUs NVIDIA y han sido optimizados para una versión antigua de CUDA.

\subsection{Algoritmos de olvido de caché}
Los algoritmos sin caché \cite{DBLP:conf/focs/FrigoLPR99} han atraído una atención considerable, ya que son portables a casi todos los entornos y arquitecturas. Se han propuesto algoritmos y estructuras de datos para tareas básicas como la ordenación, la búsqueda o el procesamiento de consultas \cite{DBLP:conf/sigmod/HeLLY07} y para tareas especializadas como la reordenación de rayos \cite{DBLP:journals/tog/MoonBKCKBNY10} o la búsqueda de homología en bioinformática \cite{DBLP:journals/bmcbi/FerreiraRR14}. Dos conceptos algorítmicos importantes de los algoritmos que no dependen de la caché son el acceso localizado a la memoria y el divide y vencerás. La curva de Hilbert integra ambas ideas. La curva de Hilbert define un ordenamiento 1D de los puntos de un espacio bidimensional de tal manera que cada punto es visitado una vez. Bader et al. propusieron utilizar la curva de Peano para la multiplicación de matrices y la descomposición LU \cite{DBLP:conf/para/BaderM06, DBLP:conf/europar/Bader08}. Los algoritmos procesan las matrices de entrada de forma recursiva y por bloques, donde la curva de Peano guía el orden de procesamiento y, por tanto, el patrón de acceso a la memoria. En \cite{loopsjournal}, se han aplicado bucles con memoria caché a la agrupación de K-means y a la multiplicación de matrices. 

\subsection{Técnicas optimizadas para tareas específicas o hardware}
\noindent La biblioteca BLAS (Basic Linear Algebra Subprograms) \cite{DBLP:journals/toms/DongarraCHD90} proporciona operaciones básicas de álgebra lineal junto con interfaces de programación para C y Fortran. BLAS está altamente optimizado para el hardware: existen implementaciones específicas para diversas infraestructuras, por ejemplo, ACML para procesadores AMD Opteron o CUBLAS para GPUs NVIDIA. La Math Kernel Library (MKL) contiene rutinas de procesamiento matemático altamente vectorizadas para los procesadores Intel. Estas implementaciones son muy específicas del hardware y, en su mayoría, están optimizadas por el proveedor. Además, están diseñadas para soportar eficazmente operaciones específicas de álgebra lineal. Los experimentos demuestran que el enfoque sin caché alcanza un rendimiento mejor que BLAS en la tarea de la unión por similitud para puntos de dimensiones en el rango de $\{2,...,64\}$.

\section{Aplicaciones}
Estas técnicas se investigan para los algoritmos de DM en las siguientes aplicaciones:
\begin{itemize}
\item Electrocardiografía (ECG, 1D),
\item Electroencefalografía (EEG, 1D),
\item Tomografía computarizada (CT, 3D, 4D),
\item SPECT (4D),
\item Resonancia Magnética Estructural (MRI, 3D),
\item Resonancia Magnética Funcional (fMRI, 4D),
\item Imagen de Tensor de Difusión (DTI, 6D),
\item Ultrasonido (US, 2D),
\item Microscopía, Dermascopía (2D),
\item, etc.
\end{itemize}

\section{Conclusión}
Este estudio revisa varios enfoques de DM de muchos grupos de investigación en todo el mundo. El hardware informático moderno permite el desarrollo de aplicaciones de alto rendimiento para el análisis de datos en muchos niveles diferentes. La atención se centra en los modernos procesadores multinúcleo incorporados a los ordenadores básicos actuales, que suelen encontrarse en los institutos universitarios como pequeños servidores y ordenadores de estación de trabajo. Por tanto, no son deliberadamente ordenadores de alto rendimiento. Los procesadores multinúcleo modernos constan de varios (de 2 a más de 100) núcleos informáticos, que funcionan de forma independiente entre sí según el principio de ``instrucciones múltiples, datos múltiples'' (MIMD). Tienen una memoria principal común (memoria compartida). Cada uno de estos núcleos informáticos dispone de varias (2-16) unidades aritméticas-lógicas, que pueden realizar simultáneamente la misma operación aritmética sobre varios datos de forma vectorial (single instruction multiple data, SIMD). Los algoritmos de DM deben utilizar ambos tipos de paralelismo (SIMD y MIMD), siendo el acceso a la memoria principal (componente centralizado) la principal barrera para aumentar la eficiencia. Investigamos estos problemas de rendimiento en el contexto de aplicaciones de la industria sanitaria como ECG, EEG, CT, SPECT, fMRI, DTI, ultrasonidos, microscopía, dermascopía, etc.

\nocite{DBLP:conf/sigmod/BohmFP08,DBLP:conf/icdt/BerchtoldBKK01,DBLP:conf/icdm/BohmK02,10.1007/BFb0000120,DBLP:conf/icde/BohmOPY07,
DBLP:conf/cikm/BohmBBK00,DBLP:journals/jiis/BohmBKM00,DBLP:conf/adl/BohmBKS00,DBLP:conf/edbt/BohmK00,
DBLP:journals/sadm/AchtertBDKZ08,DBLP:journals/bioinformatics/BaumgartnerBBMWOLR04,DBLP:journals/jbi/BaumgartnerBB05,
DBLP:conf/icdt/BerchtoldBKK01,DBLP:conf/edbt/BohmP08,DBLP:conf/kdd/AchtertBKKZ06,DBLP:conf/cikm/BohmFOPW09,
DBLP:conf/ssdbm/BohmPS06,DBLP:conf/kdd/BohmHMP09,DBLP:conf/dawak/BerchtoldBKKX00,DBLP:conf/sdm/AchtertBDKZ08,
DBLP:journals/jdi/BaumgartnerGBF05,DBLP:journals/kais/MaiHFPB15,DBLP:journals/tlsdkcs/BohmNPWZ09,
DBLP:conf/dexa/BohmK03,DBLP:conf/icde/BohmGKPS07,DBLP:conf/miccai/DyrbaEWKPOMPBFFHKHKT12,DBLP:conf/kdd/PlantB11,
DBLP:journals/bioinformatics/PlantBTB06,DBLP:conf/icdm/ShaoPYB11,DBLP:conf/icdm/PlantWZ09,
DBLP:journals/tkdd/BohmFPP07,DBLP:conf/pakdd/BohmGOPPW10,DBLP:conf/btw/BohmNPZ09,DBLP:conf/icdm/MaiGP12,
DBLP:journals/kais/ShaoWYPB17,DBLP:conf/kdd/YeGPB16,DBLP:conf/kdd/FengHKBP12,DBLP:journals/envsoft/YangSSBP12,
DBLP:conf/icdm/GoeblHPB14,DBLP:conf/kdd/AltinigneliPB13,DBLP:conf/icdm/YeMHP16,DBLP:conf/kdd/Plant12,
DBLP:conf/kdd/Plant12,DBLP:conf/cikm/BohmBBK00,DBLP:journals/kais/BohmK04,DBLP:conf/icdm/BohmK02,
DBLP:conf/icdt/BerchtoldBKK01,loopsjournal,Bially1969SpacefillingCT,
Prusinkiewicz:1986:GAL:16564.16608,10.1007/BFb0000120,DBLP:conf/icdm/BohmK02,
DBLP:conf/ssdbm/AchtertBKKZ07,DBLP:conf/pkdd/AchtertBKKMZ06,SHAO20122756,DBLP:journals/tkde/ShaoHBYP13,DBLP:conf/icdm/BohmK02,
DBLP:conf/ssdbm/AchtertBKZ06,DBLP:conf/cikm/BohmBBK00,DBLP:journals/jiis/BohmBKM00}
\bibliographystyle{alpha}
\bibliography{bibliography, bib-new}
\end{document}